\documentclass[pdflatex,sn-mathphys-num]{sn-jnl}

\usepackage{graphicx}%
\usepackage{multirow}%
\usepackage{amsmath,amssymb,amsfonts}%
\usepackage{amsthm}%
\usepackage{mathrsfs}%
\usepackage[title]{appendix}%
\usepackage{xcolor}%
\usepackage{textcomp}%
\usepackage{manyfoot}%
\usepackage{booktabs}%
\usepackage{algorithm}%
\usepackage{algorithmicx}%
\usepackage{algpseudocode}%
\usepackage{listings}%
\usepackage{bm}

\theoremstyle{thmstyleone}%
\theoremstyle{thmstyletwo}%

\theoremstyle{thmstylethree}%

\begin{document}

\title[Article Title]{BoundAD: Boundary-Aware Negative Generation for Time Series Anomaly Detection}



\author[1]{\fnm{Xiancheng} \sur{Wang}}\email{24S130294@hit.std.edu.cn}
\author*[1]{\fnm{Lin} \sur{Wang}}\email{wanglin\_007@hitwh.edu.cn}

\author[1]{\fnm{Rui} \sur{Wang}}\email{wangrui@hitwh.edu.cn}
\author[2]{\fnm{Zhibo} \sur{Zhang}}\email{zhangzhibo@cqsf.com}
\author[1]{\fnm{Minghang} \sur{Zhao}}\email{zhaomh@hit.edu.cn}

\affil*[1]{\orgdiv{School of Ocean Engineering}, \orgname{Harbin Institute of Technology}, \orgaddress{\street{West Wenhua Road}, \city{Weihai}, \postcode{264209}, \state{Shandong}, \country{China}}}
\affil[2]{\orgdiv{Technical Center, Bogie Development Department}, \orgname{CRRC Qingdao Sifang Locomotive and Rolling Stock Co., Ltd.}, \orgaddress{\street{Jinhong East Road}, \city{Qingdao}, \postcode{266111}, \state{Shandong}, \country{China}}}







\abstract{Contrastive learning methods for time series anomaly detection (TSAD) heavily depend on the quality of negative sample construction. However, existing strategies based on random perturbations or pseudo-anomaly injection often struggle to simultaneously preserve temporal semantic consistency and provide effective decision-boundary supervision. Most existing methods rely on prior anomaly injection, while overlooking the potential of generating hard negatives near the data manifold boundary directly from normal samples themselves. To address this issue, we propose a reconstruction-driven boundary negative generation framework that automatically constructs hard negatives through the reconstruction process of normal samples. Specifically, the method first employs a reconstruction network to capture normal temporal patterns, and then introduces a reinforcement learning strategy to adaptively adjust the optimization update magnitude according to the current reconstruction state. In this way, boundary-shifted samples close to the normal data manifold can be induced along the reconstruction trajectory and further used for subsequent contrastive representation learning. Unlike existing methods that depend on explicit anomaly injection, the proposed framework does not require predefined anomaly patterns, but instead mines more challenging boundary negatives from the model's own learning dynamics. Experimental results show that the proposed method effectively improves anomaly representation learning and achieves competitive detection performance on the current dataset.}


\maketitle

\section{Introduction}\label{Introduction}

Time series anomaly detection (TSAD) plays an important role in a wide range of real-world applications, including industrial monitoring, cloud system operations, Internet of Things (IoT) sensing, healthcare, and cybersecurity, where anomalies often correspond to equipment faults, system instability, external attacks, or other potential risk events \cite{DeepTSADSurvey}. In recent years, with the rapid growth of high-dimensional, multivariate, and strongly nonlinear time series data, deep learning-based TSAD methods have attracted increasing attention and gradually become one of the dominant technical paradigms in this field \cite{DeepTSADSurvey,OmniAnomaly,TranAD}.

Existing TSAD methods can be roughly categorized into three groups: reconstruction-based, forecasting-based, and representation learning-based approaches. Reconstruction-based methods characterize normal patterns and detect anomalies through reconstruction errors, with representative examples including models based on stochastic recurrent networks or variational autoencoders \cite{OmniAnomaly}. Forecasting-based or temporal modeling methods instead identify anomalies by measuring deviations from predicted future patterns. In addition, a growing number of recent studies have enhanced temporal modeling capability through Transformers, hierarchical one-class classification, and attention mechanisms, such as THOC, Anomaly Transformer, and TranAD \cite{THOC,AnomalyTransformer,TranAD}. Although these methods have achieved promising progress on multiple benchmarks, under the unsupervised setting they still mainly rely on one-class modeling of normal patterns. As a result, they may learn overly compact normal boundaries, which often leads to high false positive rates when normal fluctuations overlap with anomalous perturbations in the feature space \cite{COUTA}.

To alleviate the difficulty caused by the lack of labeled supervision, self-supervised representation learning, especially contrastive learning, has recently been introduced into TSAD and shown promising potential \cite{TS2Vec,DCdetector}. The core idea of contrastive learning is to pull semantically similar samples closer while pushing semantically dissimilar samples farther apart in the representation space, a principle that has been widely validated in computer vision and sequential representation learning \cite{SimCLR,CPC}. In the time series domain, TS2Vec learns universal temporal representations through hierarchical contrastive objectives, while DCdetector further improves anomaly discrimination with dual attention and purely contrastive objectives \cite{TS2Vec,DCdetector}. However, most existing contrastive TSAD methods largely inherit the construction paradigm from computer vision, where augmented views are treated as positive samples and temporally distant windows are treated as negative samples. For time series, such assumptions do not always hold. On the one hand, certain augmentation operations may destroy anomaly-sensitive local semantics, causing the so-called positive samples to deviate from the original normal pattern. On the other hand, temporally distant segments may still belong to the same normal operating state, which introduces semantic conflicts into negative sample construction \cite{TS2Vec,DCdetector}.

Recent studies have further shown that merely modeling normality is often insufficient for learning robust anomaly boundaries, and that appropriately incorporating anomaly-aware information can effectively improve the discriminability of the representation space. For example, COUTA alleviates data contamination and the lack of anomaly priors through calibrated one-class learning, while Outlier Exposure demonstrates that explicitly introducing auxiliary out-of-distribution or abnormal samples can help shape more reliable decision boundaries \cite{COUTA,OutlierExposure}. These studies suggest that, for unsupervised TSAD, the key is not only to learn what is normal, but also to construct anomaly-oriented signals that are genuinely useful for boundary learning in the absence of true anomaly labels.

Meanwhile, reinforcement learning (RL) provides a new perspective on this problem. RL is well suited for sequential decision-making and adaptive policy optimization, and has recently been applied to anomaly detection, active querying, and adaptive temporal analysis \cite{RLAD,ADT,MetaAAD}. Unlike fixed perturbation rules or static heuristic injection, RL can dynamically adjust actions according to the current state and historical feedback, thereby gradually learning more suitable strategies for sampling, control, or generation \cite{RLAD,ADT}. For contrastive TSAD, the quality and difficulty of negative samples directly affect the separability of the learned representation space, and existing studies have also shown that properly designed hard negatives are often more effective than random negatives for improving discriminative representations \cite{HardNegativeCL}. This naturally raises an important question: can reinforcement learning be used to adaptively control the generation process of negative samples, so that they no longer rely on fixed augmentation rules but instead progressively move toward more informative boundary regions?

Motivated by the above observations, we propose \textbf{BoundTSAD}, a reinforcement learning-guided self-supervised contrastive framework for time series anomaly detection. The core idea of the proposed method is to introduce an adaptive reinforcement learning policy to control the negative sample generation process in the absence of anomaly labels, enabling the model to construct contrastive negatives located near the boundary of normality and with stronger discriminative value. In contrast to existing methods that either rely solely on one-class normality modeling or are limited to fixed anomaly injection heuristics, our method attempts to organically combine boundary exploration with contrastive representation learning, thereby reducing false alarms under subtle anomaly scenarios and improving the generalization capability of the model.

Specifically, the main contributions of this paper are summarized as follows:
\begin{itemize}
    \item We propose \textbf{BoundTSAD}, a reinforcement learning-guided self-supervised contrastive framework for unsupervised time series anomaly detection, which integrates boundary-aware negative generation with representation learning.
    \item We design a \textbf{boundary-aware negative generation mechanism} that adaptively controls the negative sample generation process through policy learning, producing training signals that are better aligned with anomaly boundary learning.
    \item We develop a \textbf{contrastive training strategy based on boundary hard negatives}, which improves the separability between normal and anomalous patterns in the latent space and enhances the detection of fine-grained anomalies.
    \item We conduct a preliminary feasibility study and empirical evaluation, and the results show that the proposed method achieves promising performance on the current dataset.
\end{itemize}

\begin{figure}
    \centering
    \includegraphics[width=1\linewidth]{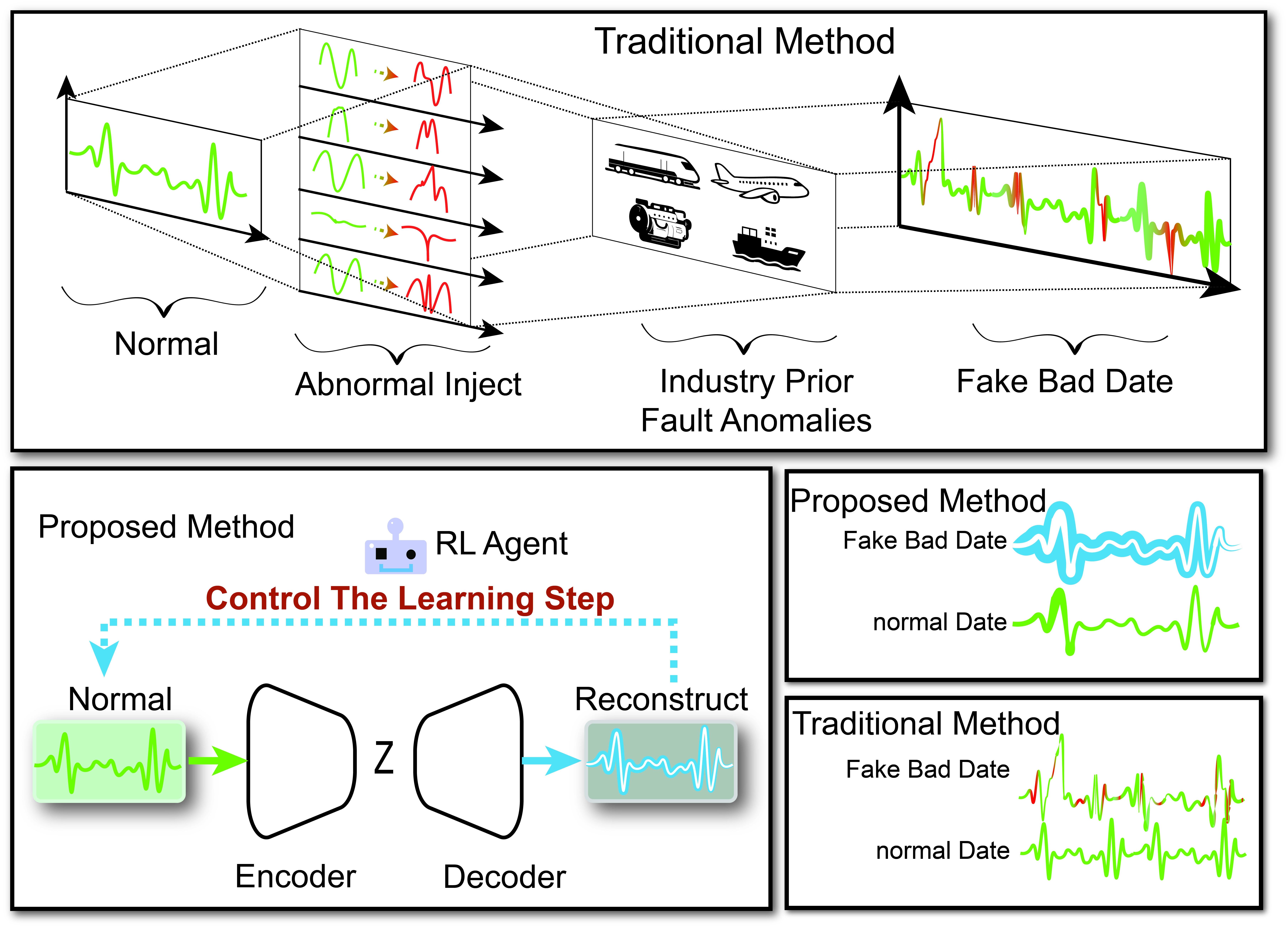}
    \caption{The difference of the methods}
    \label{fig:conbat}
\end{figure}

\section{Method}
\begin{figure}
    \centering
    \includegraphics[width=1\linewidth]{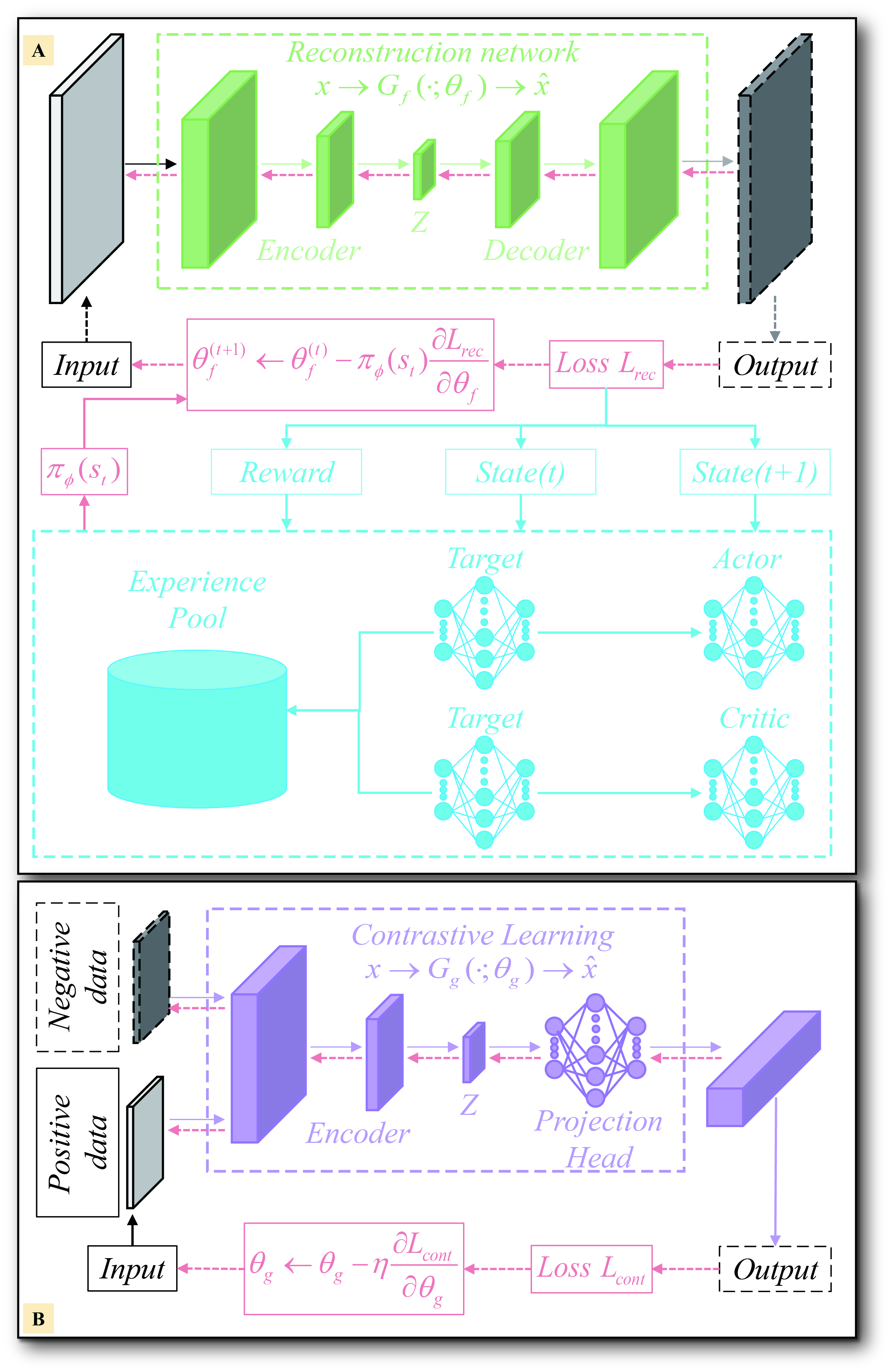}
    \caption{The structure of the model }
    \label{fig:placeholder}
\end{figure}
This paper proposes a three-stage framework for time series anomaly detection, consisting of reconstruction pretraining, reinforcement learning-guided pseudo-anomaly generation with contrastive representation learning, and prototype-driven discriminative refinement. The core objective of the proposed method is not merely to improve a single evaluation metric, but to learn an anomaly scoring function that keeps the scores of normal samples consistently close to low values while rapidly increasing when anomalies occur.

Let the original time series be denoted as
\begin{equation}
\bm{x} = \{x_1, x_2, \dots, x_T\},
\end{equation}
where $T$ is the sequence length. A sliding window of length $w$ is applied to construct window-level samples:
\begin{equation}
\bm{X}_i = [x_i, x_{i+1}, \dots, x_{i+w-1}]^\top, \quad i = 1,2,\dots,T-w+1.
\end{equation}
The goal is to learn an anomaly scoring function $s(\bm{X}_i)$ such that normal windows receive low scores while anomalous windows receive high scores.

\subsection{Overall Framework}

The proposed method consists of the following three stages:

\begin{enumerate}
    \item \textbf{Stage 1: Reconstruction model pretraining.} A reconstruction model is trained on normal temporal patterns to obtain basic temporal representation and reconstruction ability.
    \item \textbf{Stage 2: Reinforcement learning-guided pseudo-anomaly generation and triplet representation learning.} Based on the pretrained reconstruction model, reinforcement learning is introduced to adaptively control the parameter update magnitude and generate pseudo-anomalous windows. The normal and pseudo-anomalous windows are then used to train a triplet encoder, yielding a more discriminative embedding space.
    \item \textbf{Stage 3: Prototype-based discriminative refinement.} On top of the embedding space learned in Stage 2, a set of learnable prototypes is introduced. By jointly optimizing normal compactness, anomaly separation, prototype dispersion, and prototype balancing constraints, the normal manifold is further refined, and the distance from a sample to its nearest prototype is used as the anomaly score.
\end{enumerate}

It should be emphasized that Stage 3 is not a simple KMeans-based post-processing step, but rather a \textbf{trainable prototype neural discrimination module}. This stage inherits the encoder parameters learned in Stage 2 and continues to jointly optimize both the encoder and prototype parameters through backpropagation. Therefore, it is essentially a neural network-driven discriminative refinement process rather than a static clustering procedure.

\subsection{Stage 1: Reconstruction Model Pretraining}

We first construct a reconstruction model $M_{\theta}$, whose backbone consists of a linear input mapping layer, an ExtremeKAN nonlinear enhancement module, positional encoding, and a Transformer encoder. Given an input window $\bm{X}_i$, the model outputs its reconstruction:
\begin{equation}
\hat{\bm{X}}_i = M_{\theta}(\bm{X}_i).
\end{equation}

The training objective is to minimize the window reconstruction error:
\begin{equation}
\mathcal{L}_{rec} = \frac{1}{N}\sum_{i=1}^{N}\|\hat{\bm{X}}_i - \bm{X}_i\|_2^2,
\end{equation}
where $N$ is the number of training windows. After this stage, the model acquires a basic ability to model normal temporal structures, providing initialization parameters for the subsequent reinforcement learning-based step control and pseudo-anomaly generation.

\subsection{Stage 2: Reinforcement Learning-Guided Pseudo-Anomaly Generation}

\subsubsection{State Construction and Action Output}

In Stage 2, the reconstruction error is not directly used for anomaly detection. Instead, it serves as a control signal for reinforcement learning to generate more discriminative pseudo-anomalous samples. For each batch of normal windows, the current reconstruction loss is first computed, and a state vector is constructed as
\begin{equation}
\bm{s}_t = [l_{t-k+1}, l_{t-k+2}, \dots, l_t, u_{t-1}],
\end{equation}
where $l_t$ denotes the reconstruction loss at time step $t$, $k$ is the length of the historical loss sequence, and $u_{t-1}$ is the signed update applied at the previous step.

The Actor network then outputs an action $a_t$ according to the current state $\bm{s}_t$:
\begin{equation}
a_t = \pi_{\phi}(\bm{s}_t), \quad a_t \in [a_{min}, a_{max}],
\end{equation}
where $\pi_{\phi}$ denotes the policy network parameterized by $\phi$. In implementation, the Actor is realized as a multi-layer perceptron, making this part a standard neural controller.

\subsubsection{Signed Step Mapping}

To make the action directly applicable to the reconstruction model, the action is mapped to a signed step $u_t$. Let the current reconstruction loss be $l_t$, and let $L_{low}$ and $L_{up}$ denote the predefined lower and upper bounds, respectively. Then,
\begin{equation}
u_t =
\begin{cases}
\eta_{pos} \cdot a_t, & l_t > L_{up}, \\
-\eta_{neg} \cdot a_t, & l_t < L_{low}, \\
0, & L_{low} \le l_t \le L_{up},
\end{cases}
\end{equation}
where $\eta_{pos}$ and $\eta_{neg}$ denote the base step sizes for positive and negative updates, respectively.

The intuition behind this design is as follows: when the reconstruction error is too high, the model is encouraged to update in the direction that reduces the error; when the error is too low, the model parameters are perturbed in the reverse direction to construct pseudo-anomalous windows that deviate from the normal manifold; and when the error lies within the target band, no update is applied.

\subsubsection{Reward Design}

Let the losses before and after the update be denoted by $l_t^{before}$ and $l_t^{after}$, respectively, and let $l^\ast$ be the target loss. The reward is defined as
\begin{equation}
r_t = |l_t^{before} - l^\ast| - |l_t^{after} - l^\ast| + \mathbb{I}(L_{low} \le l_t^{after} \le L_{up}),
\end{equation}
where $\mathbb{I}(\cdot)$ is the indicator function. The reward increases when the updated loss moves closer to the target interval, and an additional reward is assigned when the updated loss falls into the desired band.

\subsubsection{Pseudo-Anomalous Window Construction}

After obtaining the action, a single gradient-based manual parameter update is applied to the reconstruction model:
\begin{equation}
\theta' = \theta - u_t \nabla_{\theta}\mathcal{L}_{rec}.
\end{equation}
The updated model is then used to generate the reconstructed window
\begin{equation}
\bm{X}_i^{neg} = M_{\theta'}(\bm{X}_i^{pos}),
\end{equation}
where $\bm{X}_i^{pos}$ denotes a normal window and $\bm{X}_i^{neg}$ denotes its corresponding pseudo-anomalous window. In this way, a normal sample set $\mathcal{P}$ and a pseudo-anomaly set $\mathcal{N}$ are obtained for subsequent representation learning.

\subsection{Stage 2: Triplet Contrastive Representation Learning}

To further enlarge the difference between normal and anomaly-related samples in the representation space, the window encoder $f(\cdot)$ is trained using the positive and negative samples generated in Stage 2. For each normal window $\bm{X}^{pos}$, a lightly augmented version $\tilde{\bm{X}}^{pos}$ is first constructed as the positive pair. The triplet is then formed as
\begin{equation}
\bm{z}_a = f(\bm{X}^{pos}), \quad
\bm{z}_p = f(\tilde{\bm{X}}^{pos}), \quad
\bm{z}_n = f(\bm{X}^{neg}).
\end{equation}

The triplet loss is defined as
\begin{equation}
\mathcal{L}_{tri} = \frac{1}{N}\sum_{i=1}^{N}\max\left(0, \|\bm{z}_a^{(i)}-\bm{z}_p^{(i)}\|_2 - \|\bm{z}_a^{(i)}-\bm{z}_n^{(i)}\|_2 + m\right),
\end{equation}
where $m$ is the margin parameter.

To further make the representations of normal samples before and after augmentation more compact, a compactness loss is introduced:
\begin{equation}
\mathcal{L}_{com} = \frac{1}{N}\sum_{i=1}^{N}\|\bm{z}_a^{(i)}-\bm{z}_p^{(i)}\|_2^2.
\end{equation}

Therefore, the overall representation learning objective in Stage 2 is
\begin{equation}
\mathcal{L}_{stage2} = \mathcal{L}_{tri} + \lambda \mathcal{L}_{com},
\end{equation}
where $\lambda$ is a weighting coefficient. After this stage, the encoder $f(\cdot)$ is expected to map normal windows, augmented normal windows, and RL-generated pseudo-anomalous windows into a more discriminative embedding space.

\subsection{Stage 3: Prototype-Based Discriminative Refinement}

\subsubsection{Whether Stage 3 Uses a Neural Network}

It should be explicitly clarified here that \textbf{Stage 3 does use a neural network}.

This is because Stage 3 does not simply perform a fixed clustering on the embeddings produced by Stage 2. Instead, it constructs a trainable model containing two sets of parameters:
\begin{equation}
G = \{f_{\psi}(\cdot), \bm{C}\},
\end{equation}
where $f_{\psi}(\cdot)$ denotes the encoder inherited from Stage 2, and $\bm{C}=\{\bm{c}_1,\bm{c}_2,\dots,\bm{c}_K\}$ denotes the set of $K$ learnable prototype vectors. During training, the encoder parameters $\psi$ and the prototype parameters $\bm{C}$ are jointly updated through gradient backpropagation. Therefore, Stage 3 is essentially a \textbf{neural discriminative model composed of an encoder and learnable prototype parameters}.

\subsubsection{Prototype Initialization and Soft Assignment}

At the beginning of Stage 3, the encoder trained in Stage 2 is first used to encode normal windows, and several normal embeddings are selected to initialize the prototype centers. For any sample embedding $\bm{z}$, the squared distance to the $k$-th prototype is defined as
\begin{equation}
d_k(\bm{z}) = \|\bm{z} - \bm{c}_k\|_2^2.
\end{equation}

For normal samples, the soft assignment weight is further defined as
\begin{equation}
\alpha_k(\bm{z}) = \frac{\exp\left(-d_k(\bm{z})/\tau\right)}{\sum_{j=1}^{K}\exp\left(-d_j(\bm{z})/\tau\right)},
\end{equation}
where $\tau$ is the temperature coefficient. The introduction of soft assignment prevents each normal sample from being rigidly assigned to only one prototype and instead allows it to contribute probabilistically to multiple prototypes, thereby improving training stability.

\subsubsection{Normal Compactness Loss}

For the normal window set $\mathcal{P}$, the embeddings are expected to stay close to some prototypes. Thus, the normal compactness loss is defined as
\begin{equation}
\mathcal{L}_{n} =
\frac{1}{|\mathcal{P}|}
\sum_{\bm{z}\in \mathcal{P}}
\sum_{k=1}^{K}
\alpha_k(\bm{z}) d_k(\bm{z}).
\end{equation}
This term encourages normal samples to form compact cluster structures around the prototypes.

\subsubsection{Anomaly Separation Loss}

For the pseudo-anomalous window set $\mathcal{N}$, the embeddings are expected to stay far away from all normal prototypes. The anomaly score is first defined as the distance to the nearest prototype:
\begin{equation}
s(\bm{z}) = \min_{k} d_k(\bm{z}).
\end{equation}
The anomaly separation loss is then formulated as
\begin{equation}
\mathcal{L}_{a} =
\frac{1}{|\mathcal{N}|}
\sum_{\bm{z}\in \mathcal{N}}
\max(0, m_p - s(\bm{z})),
\end{equation}
where $m_p$ is the Stage 3 margin parameter. This term requires pseudo-anomalies to maintain at least a certain distance from the normal prototypes.

\subsubsection{Prototype Dispersion Loss}

If multiple prototypes are too close to one another, they may degenerate into only a few centers, thereby weakening the representation of multimodal normal structures. Therefore, the prototype dispersion loss is defined as
\begin{equation}
\mathcal{L}_{sep} =
\frac{1}{K(K-1)}
\sum_{i \ne j}
\max(0, m_p - \|\bm{c}_i - \bm{c}_j\|_2)^2.
\end{equation}
This term forces different prototypes to maintain sufficient separation so that they can cover different modes of normal samples.

\subsubsection{Prototype Balancing Loss}

If, during training, most normal samples are assigned to only a few prototypes, prototype utilization becomes highly imbalanced. To address this issue, the average usage of each prototype within a batch is computed as
\begin{equation}
u_k = \frac{1}{|\mathcal{P}|}\sum_{\bm{z}\in \mathcal{P}} \alpha_k(\bm{z}),
\end{equation}
and the target distribution is set to the uniform distribution
\begin{equation}
\bar{u}_k = \frac{1}{K}.
\end{equation}
The balancing loss is then defined as
\begin{equation}
\mathcal{L}_{bal} = \mathrm{KL}(\bm{u}\,\|\,\bar{\bm{u}}).
\end{equation}
This term alleviates prototype collapse and improves the modeling capability of multiple prototypes.

\subsubsection{Overall Loss of Stage 3}

Finally, the total loss of Stage 3 is defined as
\begin{equation}
\mathcal{L}_{stage3}
=
\alpha \mathcal{L}_{n}
+ \beta \mathcal{L}_{a}
+ \gamma \mathcal{L}_{sep}
+ \delta \mathcal{L}_{bal},
\end{equation}
where $\alpha,\beta,\gamma,\delta$ are the weighting coefficients of each loss term. During training, the encoder parameters and the prototype parameters are jointly updated through backpropagation, and prototype normalization is performed after each optimization step to maintain the stability of the representation space.

\subsection{Anomaly Scoring and Decision Rule}

During testing, for any input window $\bm{X}_i$, the Stage 3 model first outputs its embedding representation:
\begin{equation}
\bm{z}_i = f_{\psi}(\bm{X}_i),
\end{equation}
and then computes its distances to all prototypes. The minimum distance is taken as the anomaly score:
\begin{equation}
s_i = \min_{k}\|\bm{z}_i - \bm{c}_k\|_2^2.
\end{equation}

Given a threshold $\tau_s$, the predicted label is defined as
\begin{equation}
\hat{y}_i =
\begin{cases}
1, & s_i \ge \tau_s, \\
0, & s_i < \tau_s.
\end{cases}
\end{equation}

Since the proposed method adopts window-level scoring, the score sequence is shifted by one window length during visualization so that the score curve is more consistent with the temporal location of the anomalous interval in an intuitive sense.

\subsection{Advantages of the Proposed Method}

Compared with traditional methods that rely only on reconstruction errors or static clustering centers, the proposed method has the following characteristics:

\begin{enumerate}
    \item Through the reinforcement learning-based step control mechanism, pseudo-anomalous samples are generated adaptively according to the current reconstruction state rather than via fixed perturbations.
    \item Through triplet representation learning, normal samples, augmented samples, and pseudo-anomalous samples form a clearer geometric structure in the embedding space.
    \item Through the learnable prototype model in Stage 3, the proposed method replaces traditional KMeans-based post-processing and enables end-to-end discriminative refinement.
    \item Through the joint constraints of normal compactness, anomaly separation, prototype dispersion, and prototype balancing, the method is better suited to characterize multimodal manifold structures under complex normal patterns.
\end{enumerate}

\subsection{Algorithm Pipeline}

\begin{algorithm}[htbp]
\caption{The proposed three-stage time series anomaly detection framework}
\begin{algorithmic}[1]
\State Train the reconstruction model $M_{\theta}$ and obtain pretrained parameters
\State Construct the sliding-window sample set
\State Initialize the RL agent, replay buffer, normal sample set, and pseudo-anomaly set
\For{each RL epoch}
    \For{each batch of normal windows}
        \State Construct the state vector
        \State Compute the current reconstruction loss
        \State The Actor outputs an action and maps it to a signed step
        \State Apply one manual parameter update to the reconstruction model
        \State Generate pseudo-anomalous windows and compute rewards
        \State Update the RL agent
        \State Collect normal windows and pseudo-anomalous windows
    \EndFor
\EndFor
\State Train the triplet encoder using normal windows, augmented normal windows, and pseudo-anomalous windows
\State Initialize the Stage 3 prototype model
\For{each Stage 3 epoch}
    \For{each batch of normal/pseudo-anomalous samples}
        \State Compute the normal compactness loss
        \State Compute the anomaly separation loss
        \State Compute the prototype dispersion loss
        \State Compute the prototype balancing loss
        \State Jointly optimize the encoder and prototype parameters
    \EndFor
\EndFor
\State Compute the distance from each test window to the nearest prototype as the anomaly score
\State Output the anomaly detection results according to the threshold
\end{algorithmic}
\end{algorithm}

\section{Experiments}
\subsection{Experimental Results Analysis}

\begin{figure}
    \centering
    \includegraphics[width=1\linewidth]{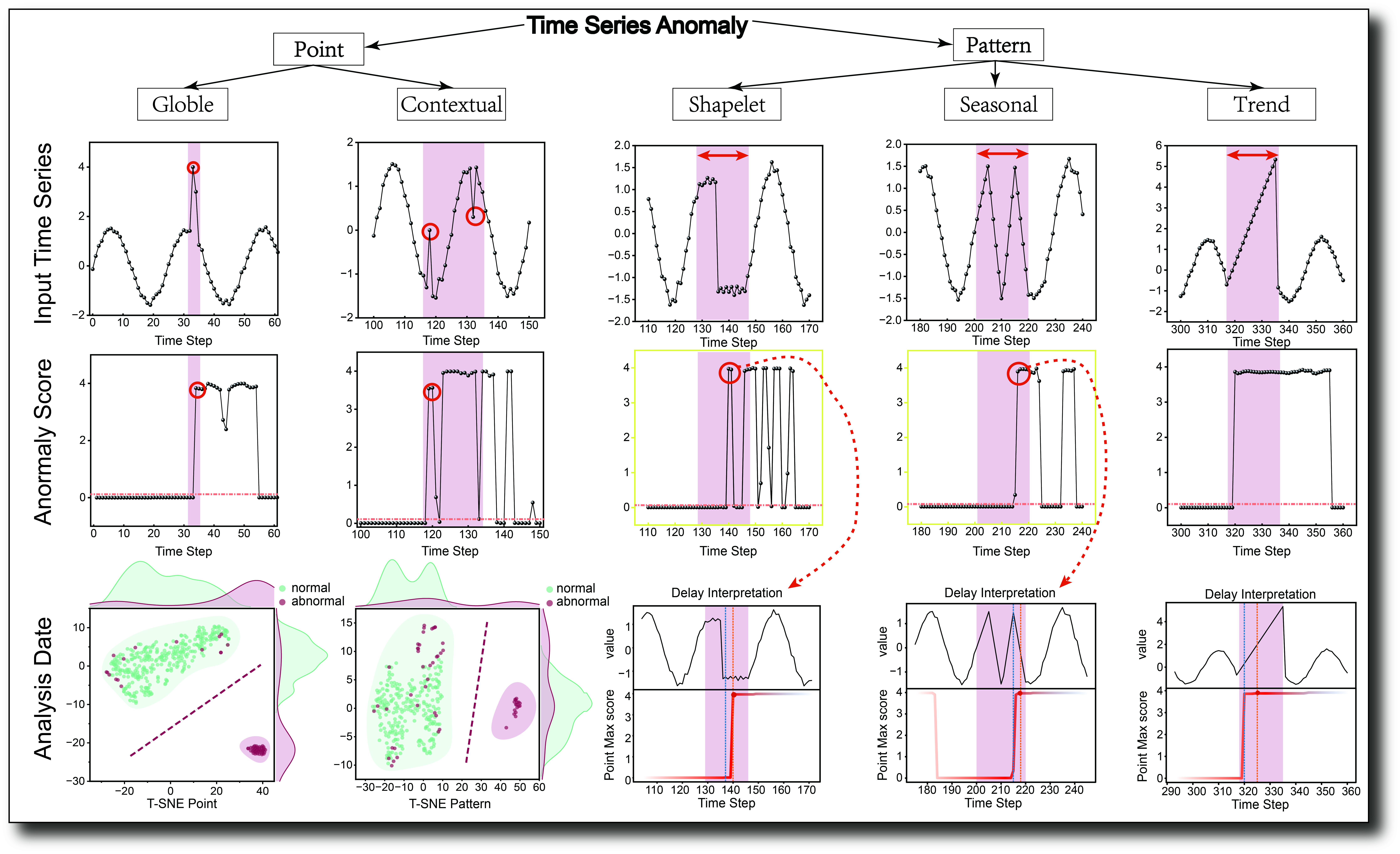}
    \caption{Display of the experience}
    \label{Display of the experience}
\end{figure}

To verify the effectiveness of the proposed three-stage framework under different anomaly types, we conducted experiments on two representative scenarios, namely \emph{Point} and \emph{Collective Seasonal}, and further analyzed the model behavior in conjunction with the visualizations in Fig.~\ref{Display of the experience}. Specifically, the two groups on the left side of the figure correspond to \emph{global} and \emph{contextual} anomalies in the point-anomaly setting, while the three groups on the right side correspond to \emph{shapelet}, \emph{seasonal}, and \emph{trend} anomalies in the pattern-anomaly setting. The first row shows the input sequence and anomalous interval, the second row shows the anomaly score curve, and the third row shows the feature distribution or delay interpretation.

\subsection{Comparative Analysis of the Two Anomaly Scenarios}

To more clearly compare the detection performance of the proposed method under different anomaly types, Table~\ref{tab:main_results} presents the quantitative results of the two experiments.

\begin{table}[htbp]
\centering
\caption{Comparison of the two experimental settings}
\label{tab:main_results}
\begin{tabular}{lcccccc}
\hline
Experiment & AUC & TP & TN & FP & FN & Predicted Anomalies \\
\hline
Seasonal& 0.9606 & 73 & 297 & 0 & 11 & 73 \\
Point & 0.9438 & 48 & 304 & 22 & 7 & 70 \\
\hline
\end{tabular}
\end{table}

As shown in Table~\ref{tab:main_results}, the proposed method achieves high ROC AUC values on both tasks, demonstrating that the three-stage framework has strong anomaly discrimination capability and a certain degree of generalization. However, the two anomaly types still exhibit noticeable differences in detection performance. Compared with the Point setting, the Collective Seasonal setting achieves a higher AUC, a lower false positive rate, and a more desirable anomaly-score separation pattern, indicating that the proposed method is better suited for modeling persistent and structured anomaly patterns. This observation is consistent with the window-level representation learning mechanism of the method itself: pattern anomalies usually induce a consistent geometric shift over a continuous temporal range, making them easier to capture jointly by the contrastive representation learning in Stage 2 and the prototype-based discrimination in Stage 3. In contrast, point anomalies only produce abrupt deviations within a very short local range, and their boundaries in the window representation space are more easily disturbed by normal fluctuations, which in turn leads to more false positives.

\subsection{Visualization Results and Interpretability Analysis}

The visualizations in Fig.~\ref{Display of the experience} further reveal the detection behavior of the model. First, in the point-anomaly setting, anomaly scores usually form sharp peaks near the anomalous locations, indicating that the model has good local localization ability. Second, in the pattern-anomaly setting, anomaly scores are more likely to exhibit sustained segment-level elevation rather than isolated point-wise spikes, suggesting that the model can extract anomaly evidence from the overall structural change within local windows.

In particular, for the three pattern anomalies, namely \emph{shapelet}, \emph{seasonal}, and \emph{trend}, the \emph{delay interpretation} shown in the bottom row indicates that the significant rise of the anomaly score usually occurs with a slight delay relative to the true anomaly onset. This is because window-level detection requires the current analysis window to accumulate sufficient structural deviation information. When the anomalous segment has not yet sufficiently entered the window, its discriminative signal remains relatively weak; once the anomalous segment begins to dominate the current window representation, the anomaly score rises rapidly. Therefore, such delay does not imply model failure, but rather is a natural consequence of the sliding-window modeling mechanism. This also suggests that the response process of the proposed method has a certain degree of interpretability.

In addition, the feature distribution shown in the figure also indicates that normal and anomalous samples exhibit a relatively clear separation trend in the embedding space. This suggests that the anomaly detection capability of the proposed method does not merely arise from thresholding at the output layer, but is instead built upon a well-structured internal representation space. This further verifies the collaborative effect of the triplet representation learning in Stage 2 and the prototype compression mechanism in Stage 3: the former enlarges the margin between normal samples and pseudo-anomalous samples, while the latter further enhances the compactness of the normal manifold and the separability of anomalous deviations.

\subsubsection{Training Process Analysis}

From the perspective of the training process, the triplet encoder in both experiments exhibits stable convergence behavior. In the Collective Seasonal experiment, the triplet loss decreases from 0.0153 to 0.0007, with a minimum value of 0.0006. In the Point experiment, the triplet loss decreases from 0.0256 to 0.0011, with a minimum value of approximately 0.0010. This indicates that, through RL-driven pseudo-anomaly construction, Stage 2 is able to effectively learn a discriminative embedding space that distinguishes normal samples, augmented samples, and pseudo-anomalous samples.

In Stage 3, the total loss in both experiments remains stable within approximately 0.045--0.050, indicating that the prototype learning process is overall well-behaved. Further observations show that the normal compactness term consistently remains at a low level, suggesting that the prototypes exert a strong attraction effect on the normal manifold. The prototype dispersion term also remains relatively stable, indicating that the prototypes maintain sufficient separation from one another. In contrast, the anomaly separation term fluctuates more noticeably in the Point setting, which is consistent with the fact that point anomalies have blurrier boundaries and are more sensitive to local noise. Overall, the pseudo-anomaly-driven representation learning in Stage 2 and the prototype discrimination mechanism in Stage 3 exhibit good synergy during training.

\subsubsection{Summary of Experimental Findings}

Combining the above quantitative results and visualization analysis, the following conclusions can be drawn. First, the proposed three-stage framework achieves strong detection performance on both Point and Collective Seasonal anomaly detection tasks, which verifies the overall effectiveness of the framework. Second, the RL-driven pseudo-anomaly construction provides contrastive representation learning with discriminative negative samples, thereby improving the separability of normal and anomalous patterns in the embedding space. Third, the prototype mechanism in Stage 3 further enhances the modeling ability of the normal manifold, making the anomaly scores exhibit a clearer geometric separation pattern. Fourth, compared with point anomalies, the proposed method performs more stably on collective and pattern anomalies, especially in terms of false alarm suppression and response to anomalous segments. These results indicate that the proposed method not only has strong detection performance, but also exhibits a certain degree of structural interpretability in its response behavior.
\section{Discussion}

\paragraph{Relation to Existing TSAD Paradigms.}
Existing time series anomaly detection methods can be broadly categorized into four groups: forecasting-based, reconstruction-based, representation-based, and hybrid methods. Forecasting-based methods usually characterize anomalies through deviations from future value prediction; reconstruction-based methods rely on the reconstructability of normal patterns; and representation-based methods focus more on enlarging the geometric boundary between normal and anomalous samples in the latent space. In recent years, contrastive learning and one-class modeling have gradually become important trends in this direction. Overall, the proposed method can be regarded as a three-stage hybrid framework that is reconstruction-driven, representation-centered, and further refined by prototype modeling in the later stage. Specifically, Stage 1 learns the basic normal patterns through a reconstruction model; Stage 2 does not stop at the reconstruction-error level, but instead actively explores boundary samples near the normal manifold through reinforcement learning-controlled perturbation steps; Stage 3 further compresses the normal manifold and amplifies anomalous deviations by means of a prototype mechanism. Therefore, the proposed method does not simply follow the traditional paradigm of treating reconstruction error as the anomaly score, but instead attempts to unify reconstruction, pseudo-anomaly generation, contrastive representation learning, and prototype discrimination within a single detection framework.

\paragraph{Core Significance of the Proposed Method.}
From the perspective of existing literature, a long-standing difficulty in unsupervised TSAD is the absence of real anomaly supervision, while the quality of negative samples or anomaly priors often directly determines the final discriminative boundary. If the negative samples are too easy, the learned boundary becomes too loose; if they deviate too far semantically from normal patterns, they fail to provide genuinely useful near-boundary discriminative information. A key motivation of the proposed method is to avoid relying entirely on manually designed anomaly injection rules. Instead, it leverages the local generative basis provided by a pretrained reconstructor and uses reinforcement learning to control both the direction and magnitude of the signed step, so that pseudo-anomalous samples tend to appear near the normal manifold while still departing from normal patterns. The resulting pseudo-anomalies are therefore neither completely random noise nor fixed-template perturbations, but rather boundary-neighborhood samples jointly induced by the data distribution and training dynamics. For contrastive learning, such samples are more valuable than coarse negatives far away from the data manifold, because they are closer to the hard boundary that the detection task truly needs to distinguish.

\paragraph{Difference from Anomaly Injection and Calibration-Based Methods.}
The proposed method is conceptually related to existing anomaly-aware TSAD methods, but differs in emphasis. One representative line of work constructs dummy anomalies or native anomalies to calibrate the normal boundary in one-class learning; another line directly introduces anomaly injection into contrastive learning so that the model jointly learns normal representations and anomalous deviations. The common point between these approaches and ours is the recognition that merely learning from normal samples is often insufficient for forming a stable anomaly boundary. The difference lies in the fact that the proposed method does not predefine a fixed perturbation pattern for anomalies. Instead, it aims to adaptively generate boundary negatives during training through RL-controlled perturbation trajectories. In other words, existing methods are more akin to providing anomaly priors to the model, whereas our approach is closer to allowing the model to actively explore the anomaly boundary near the normal manifold. This difference makes the proposed method conceptually better suited as a boundary learning or post-training enhancement mechanism, rather than just another static anomaly injection strategy.

\paragraph{Why the Method Is More Effective for Pattern Anomalies.}
According to the current experimental observations, the proposed method is generally more stable on collective or pattern anomalies than on point anomalies, and this is consistent with the underlying mechanism of the framework. First, the discriminative unit of the method is essentially a sliding-window representation rather than an isolated single point. Therefore, it is naturally more sensitive to structural deviations that persist over a period of time. Whether the anomaly is a shapelet mutation, a seasonal phase disruption, or a trend drift, once sufficient structural discrepancy accumulates within a local window, the representation learned in Stage 2 is more likely to push it away from the normal manifold, and the prototype mechanism in Stage 3 is more likely to assign sustained high anomaly responses. In contrast, point anomalies are short-lived, highly localized, and more easily mixed with normal fluctuations, so their boundaries in the window-level latent space are inherently blurrier. This also explains why the proposed method can achieve high detection rates on point anomalies while often being less ideal in false-positive control than in pattern anomaly scenarios.

\paragraph{Understanding the Detection Delay.}
For pattern anomalies, the proposed method may sometimes exhibit a slight detection delay. This does not necessarily mean that the model fails to recognize the anomaly; rather, it is more likely a natural consequence of the window-level decision mechanism. When the anomalous segment has not yet sufficiently entered the current analysis window, its influence on the overall representation remains limited. Only when the anomalous pattern begins to dominate the window does the latent deviation and prototype distance increase rapidly. Therefore, we tend to interpret this phenomenon as a \emph{window accumulation trigger} rather than a true missed detection. This point is particularly important for industrial scenarios, where many real faults do not occur as strictly isolated points, but instead evolve gradually and reveal their statistical and structural abnormalities over time. In this sense, the proposed method has a certain interpretability advantage in modeling persistent and structural anomalies.

\paragraph{Current Limitations.}
Although the preliminary results indicate that the proposed idea is promising, the current study still has several clear limitations. First, the experimental scope remains limited, and the method has not yet been systematically evaluated on a larger, more heterogeneous, and more rigorous collection of datasets. Therefore, at the present stage, this work is better viewed as a methodological exploration rather than a fully validated general solution. Second, the stability and efficiency of the RL process in Stage 2 may still be affected by the state design, reward function, number of training rounds, and action range, and its hyperparameter sensitivity has yet to be quantified more thoroughly. Third, the number of prototypes in Stage 3, the initialization strategy, and the regularization weights may also influence the degree of normal manifold compression and the separability of anomaly scores. Fourth, the current discussion is mainly based on window-level detection, which leaves room for further improvement in anomaly localization granularity, real-time response, and early detection capability. Fifth, the method has currently been verified only on several typical anomaly types, and more sufficient evidence is still needed for cross-dataset generalization, robustness to training-set contamination, and effectiveness under complex multivariate coupling scenarios.

\paragraph{Remarks on Evaluation and the Boundary of Conclusions.}
An important recent consensus in TSAD is that dataset quality, annotation boundaries, threshold selection, and evaluation metrics can all significantly affect method conclusions. Therefore, at the current stage, the discussion of the proposed method should focus as much as possible on its mechanism rationality and preliminary effectiveness, while avoiding overly strong performance claims. In particular, for window-level detection results, reporting only a single AUC or classification statistics under a single threshold is not sufficient to fully demonstrate the merits of a method. In future work, more robust evaluation under a unified benchmark protocol is needed, such as event-level or interval-level metrics, VUS-PR, and other evaluation measures more suitable for TSAD. In addition, the stability of the conclusions should be verified under unified hyperparameter tuning and multiple random seeds.

\paragraph{Future Work.}
Based on the above analysis, there are several natural directions for future extension. First, the current RL-driven pseudo-anomaly generation mechanism can be generalized into a more universal boundary sample generation module, allowing flexible integration with forecasting-based, reconstruction-based, or foundation-model-based TSAD backbones. Second, it would be valuable to further investigate the relationship between the samples generated by the RL policy and the distribution of real anomalies, in order to analyze whether the policy learns some transferable anomaly direction or boundary perturbation pattern. Third, multi-scale windows, hierarchical prototypes, or online update mechanisms may be introduced to improve the unified modeling of short point anomalies and long-term trend anomalies. Fourth, the proposed method should be systematically evaluated on larger-scale benchmarks, especially widely used TSAD datasets such as SMD. Fifth, from a theoretical perspective, if pseudo-anomaly generation, triplet representation learning, and prototype discrimination can be further unified as a progressive approximation process of the normal manifold boundary, this would help explain the source of effectiveness of the proposed method more clearly from a geometric viewpoint.

\paragraph{Summary.}
Overall, the main value of this work lies not only in proposing a new structural combination for TSAD, but also in emphasizing a research direction that deserves further exploration: in unsupervised TSAD, the key to anomaly detection performance may not necessarily come only from stronger encoders or more complex reconstructors, but also from the active construction and utilization of hard negative samples near the normal boundary. From this perspective, the proposed method connects a reconstruction model, RL-driven boundary exploration, triplet representation learning, and prototype discrimination, providing a preliminary yet promising framework for more systematic boundary-learning-based TSAD research in the future.

\section{Conclusion}

This paper proposed a three-stage framework for unsupervised time series anomaly detection, which progressively optimizes the detection process from normal pattern modeling to anomaly boundary learning by organically combining reconstruction pretraining, RL-driven pseudo-anomaly generation, triplet representation learning, and prototype-based discrimination. Specifically, Stage 1 learns the basic representations of normal time series through a reconstruction task; Stage 2 uses reinforcement learning to control perturbation steps and generate pseudo-anomalous samples with boundary properties near the normal manifold, and then employs triplet learning to enlarge the latent-space gap between normal and pseudo-anomalous samples; Stage 3 further compresses the normal manifold and enhances anomalous deviations through a prototype mechanism, thereby producing a clearer anomaly score representation.

Experimental results show that the proposed method achieves promising detection performance on two representative anomaly detection tasks, namely Point and Collective Seasonal anomalies, and exhibits more stable false-positive control and more desirable anomaly-score separation in pattern anomaly scenarios. The visualization results further demonstrate that the proposed method can not only produce sensitive local responses to point anomalies, but also generate sustained and interpretable high-score responses to structural anomalies such as shapelet, seasonal, and trend anomalies. This indicates that the RL-driven pseudo-anomaly construction and the prototype discrimination mechanism can effectively improve boundary learning ability in unsupervised anomaly detection.

Overall, the main contribution of this work lies in proposing a three-stage paradigm that differs from the traditional practice of directly treating reconstruction error as the anomaly score. By unifying pseudo-anomaly generation, contrastive representation learning, and prototype-based discrimination within a single framework, the proposed method provides a new feasible path for hard negative construction and normal boundary modeling in unsupervised TSAD. Future work will further investigate this framework on larger-scale benchmark datasets, complex multivariate scenarios, robustness to training contamination, and multi-scale anomaly modeling, so as to further validate and improve its generality and practical value.

\section{Competing interests}
No competing interest is declared.

\section{Author contributions statement}

L.W. and X.W. conceived the main idea. L.W. and R.W. developed the detailed methodology. X.W. and Z.Z. analysed the results. X.W. and M.Z. wrote the manuscript. All authors reviewed and approved the final manuscript. 

\section*{Acknowledgments}
This work was supported by  The National Key Research and Development Program of China (2023YFB3308100), the China State Railway Group Co., Ltd. Science and Technology Research and Development Program Project (K2024J011), and the Natural Science Foundation of Shandong Province (ZR2023ME124),the National Natural Science Foundation of China (No.52475553).

\bibliography{sn-bibliography}

\end{document}